\documentclass{article}

\usepackage[dblblindworkshop, final]{neurips_2026}
\workshoptitle{AXIOM: Foundations of Efficient Deep Learning}

\usepackage[utf8]{inputenc}
\usepackage[T1]{fontenc}
\usepackage{hyperref}
\usepackage{url}
\usepackage{booktabs}
\usepackage{amsfonts}
\usepackage{nicefrac}
\usepackage{microtype}
\usepackage{xcolor}
\usepackage{amsmath}
\usepackage{graphicx}
\usepackage{subcaption}
\usepackage{float}

\title{Compute-Optimal Scaling Laws for Spiking Language Models}

\author{%
  Dan Saunders\thanks{Equal contribution.} \\
  Independent Researcher\\
  \texttt{danjsaund@gmail.com} \\
  \And
  Benjamin Ren\footnotemark[1] \\
  X School\\
  \texttt{ultraman1234502@gmail.com} \\
  \And
  Catherine He\footnotemark[1] \\
  X School\\
  \texttt{catherinezhanghe@gmail.com} \\
}

\begin{document}

\maketitle

\begin{abstract}
  Spiking neural networks target energy-efficient language modeling through sparse, binary activations, but the efficiency--accuracy tradeoff of spiking language models is uncharacterized at scale. We present the first Chinchilla-style IsoFLOP scaling study of a spiking language model, sweeping SpikeGPT-family models \citep{zhu2024spikegptgenerativepretrainedlanguage} (RWKV backbone \citep{peng2023rwkvreinventingrnnstransformer}) from width 384 to 1536 across compute budgets from $3\times10^{17}$ to $1\times10^{19}$ FLOPs on FineWeb-EDU. We report two findings. First, the compute-optimal frontier is $N_{opt} \propto C^{0.56}$, close to the $0.50$ that \citet{hoffmann2022trainingcomputeoptimallargelanguage} report for dense Transformers, so spiking does not change how parameters and data should be traded off under a compute budget. Second, against non-spiking RWKV and Transformer baselines trained the same way, the loss gap splits into a backbone cost of about $0.16$ nats that does not change with compute and a spiking cost that falls steadily with compute, from $0.34$ to $0.23$ nats over our range. These results give the first compute-optimal recipe for spiking language models and quantify the accuracy cost of their sparsity.
\end{abstract}

\section{Introduction}

Scaling laws address a basic allocation question: given a fixed compute budget $C$, how should it be split between model size $N$ and data $D$ to minimize loss? \citet{kaplan2020scalinglawsneurallanguage} established power-law scaling for Transformers, and \citet{hoffmann2022trainingcomputeoptimallargelanguage} (Chinchilla) introduced the IsoFLOP methodology and revised the optimal allocation toward roughly equal scaling of $N$ and $D$. How these laws transfer to spiking neural networks is unknown. SpikeGPT \citep{zhu2024spikegptgenerativepretrainedlanguage} binarizes activations into spikes on an RWKV backbone \citep{peng2023rwkvreinventingrnnstransformer} for energy-efficient neuromorphic inference, but was trained far below the regime where compute-optimal behavior is characterized, leaving open both how a spiking model should be scaled and what accuracy its sparsity costs.

We conduct a Chinchilla-style IsoFLOP study across a grid of compute budgets, fitting parabolas to recover compute-optimal $(N_{opt}, D_{opt})$ at each budget. Our contributions are: (1) the first compute-optimal IsoFLOP scaling law for a spiking language model, $N_{opt} \propto C^{0.56}$, close to dense Transformers; (2) non-spiking RWKV and Transformer baselines at every budget, which separate a constant backbone cost ($\sim$0.16 nats) from a spiking cost that falls steadily, from $0.34$ to $0.23$ nats; and (3) evidence that the apparent departure from the Chinchilla exponent is largely a parameter-counting artifact.

\section{Related Work}

\paragraph{Scaling Laws.}
\citet{kaplan2020scalinglawsneurallanguage} established power-law scaling of transformer loss with model size, data, and compute; \citet{hoffmann2022trainingcomputeoptimallargelanguage} (Chinchilla) revised the compute-optimal allocation toward scaling $N$ and $D$ in roughly equal proportion, using three estimators including the IsoFLOP profiles (their Approach~2) we adopt. \citet{pearce2024reconcilingkaplanchinchillascaling} show that much of the Kaplan--Chinchilla exponent discrepancy is a parameter-counting artifact (non-embedding versus total parameters), a point central to our analysis (Section~\ref{sec:convention}). \citet{shen2024scalinglawslinearcomplexity} extend IsoFLOP fitting to linear-complexity, non-softmax sequence models from 70M to 7B parameters, finding scaling broadly similar to transformers; we apply the same methodology to spiking RWKV models.

\paragraph{Spiking Language Models.}
RWKV \citep{peng2023rwkvreinventingrnnstransformer} replaces quadratic self-attention with a linear-time recurrence, removing the sequence-length term that complicates transformer FLOP accounting. \citet{zhu2024spikegptgenerativepretrainedlanguage} build SpikeGPT on this backbone, binarizing activations into discrete spikes for energy-efficient, event-driven inference on neuromorphic hardware. A separate line spikes large \emph{pretrained} LLMs post hoc \citep{xing2025spikellmscalingspikingneural}. To our knowledge, no prior work characterizes compute-optimal scaling for spiking language models trained from scratch, which we address here.

\section{Methods}

\subsection{Experimental Settings}

All models are trained with the AdamW optimizer (\(\beta_1 = 0.9\), \(\beta_2 = 0.999\)) using a batch size of 16 and a context length of 1024 tokens. Text is tokenized using the \texttt{EleutherAI/gpt-neox-20b} tokenizer \citep{black2022gptneox20bopensourceautoregressivelanguage}, and training data is drawn from the FineWeb-EDU \texttt{sample-100BT} split \citep{penedo2024finewebdatasetsdecantingweb}. Spiking neuron dynamics follow the standard leaky integrate-and-fire (LIF) model \citep{gerstner2002spiking, maass1997networks}, with the arctangent (\(\arctan\)) surrogate gradient \citep{neftci2019surrogategradientlearningspiking} used to enable backpropagation through the non-differentiable spiking threshold. The peak learning rate at each width comes from a power law $\mathrm{LR}_{\mathrm{opt}}(d) = k\,d^{-p}$ fit to a pilot sweep (Appendix~\ref{app:lr}), with cosine decay to one-tenth, so loss differences across the grid reflect the $N$/$D$ allocation rather than learning-rate tuning. We estimate training compute as $C \approx 6ND$ \citep{kaplan2020scalinglawsneurallanguage}; RWKV has no attention term, so matrix multiplies dominate.

\subsection{Parameter Counting}
\label{sec:non-embed-correction}
With a $50{,}277$-token vocabulary the input and output vocabulary matrices are $15$--$67\%$ of all parameters, so how $N$ is counted matters. We distinguish the strict non-embedding count ($13ld^2$, Kaplan), the FLOP-matmul count entering $C = 6ND$ ($N_{\backslash E} = 13ld^2 + vd$, retaining the output projection), and the total count ($N_T = 13ld^2 + 2vd$), with $v$ the vocabulary size. Following \citet{pearce2024reconcilingkaplanchinchillascaling}, total is our primary convention; Section~\ref{sec:convention} reports all three. Inverting a parameter budget to a width $d$ requires solving a cubic (Appendix~\ref{app:cardano}).

\section{Results}

\subsection{IsoFLOP profiles}
\label{sec:isoflop-profiles}

\begin{figure}[t]
    \centering
    \begin{subfigure}{0.49\textwidth}
        \includegraphics[width=\textwidth]{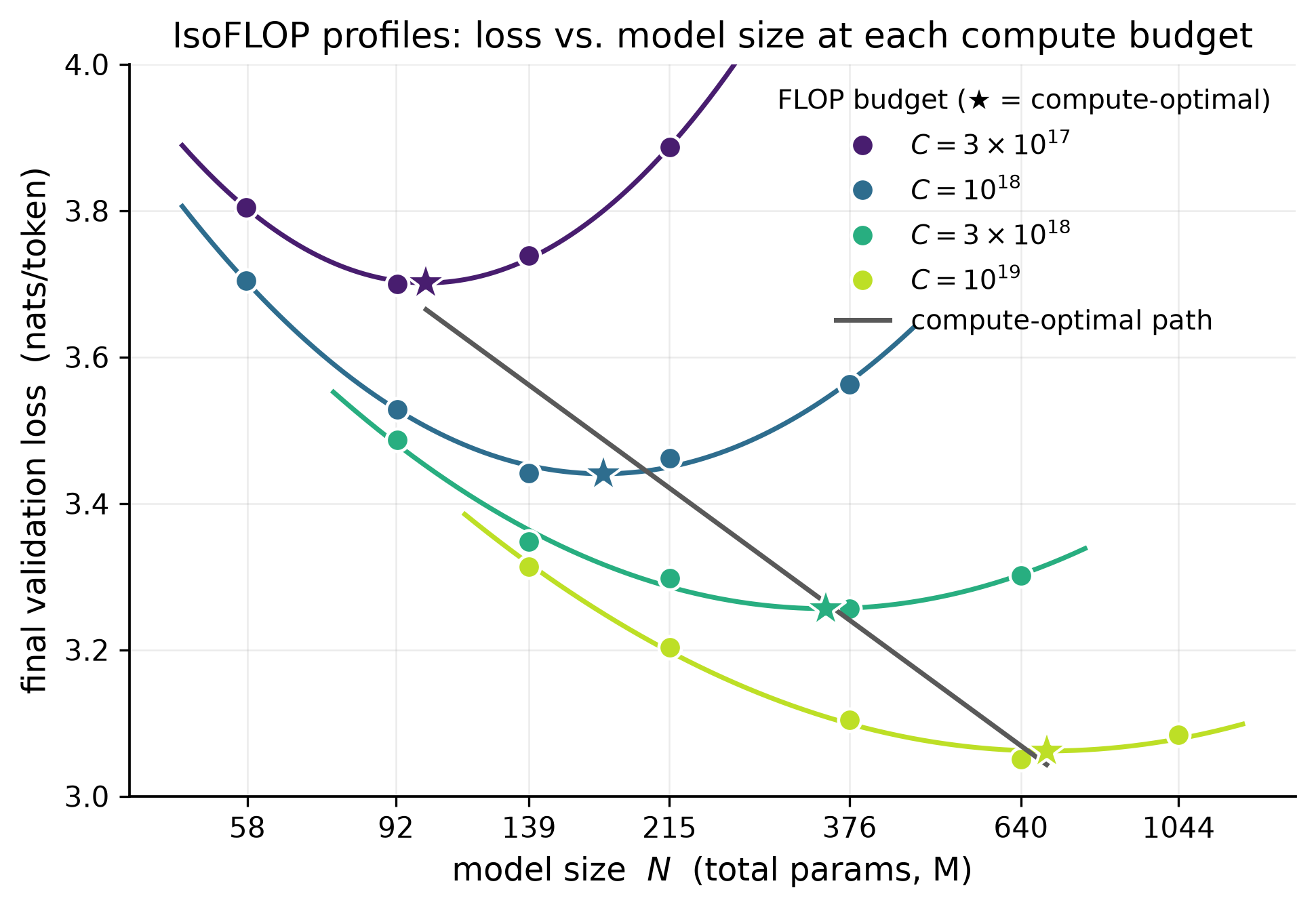}
        \caption{}\label{fig:isoflop-a}
    \end{subfigure}\hfill
    \begin{subfigure}{0.49\textwidth}
        \includegraphics[width=\textwidth]{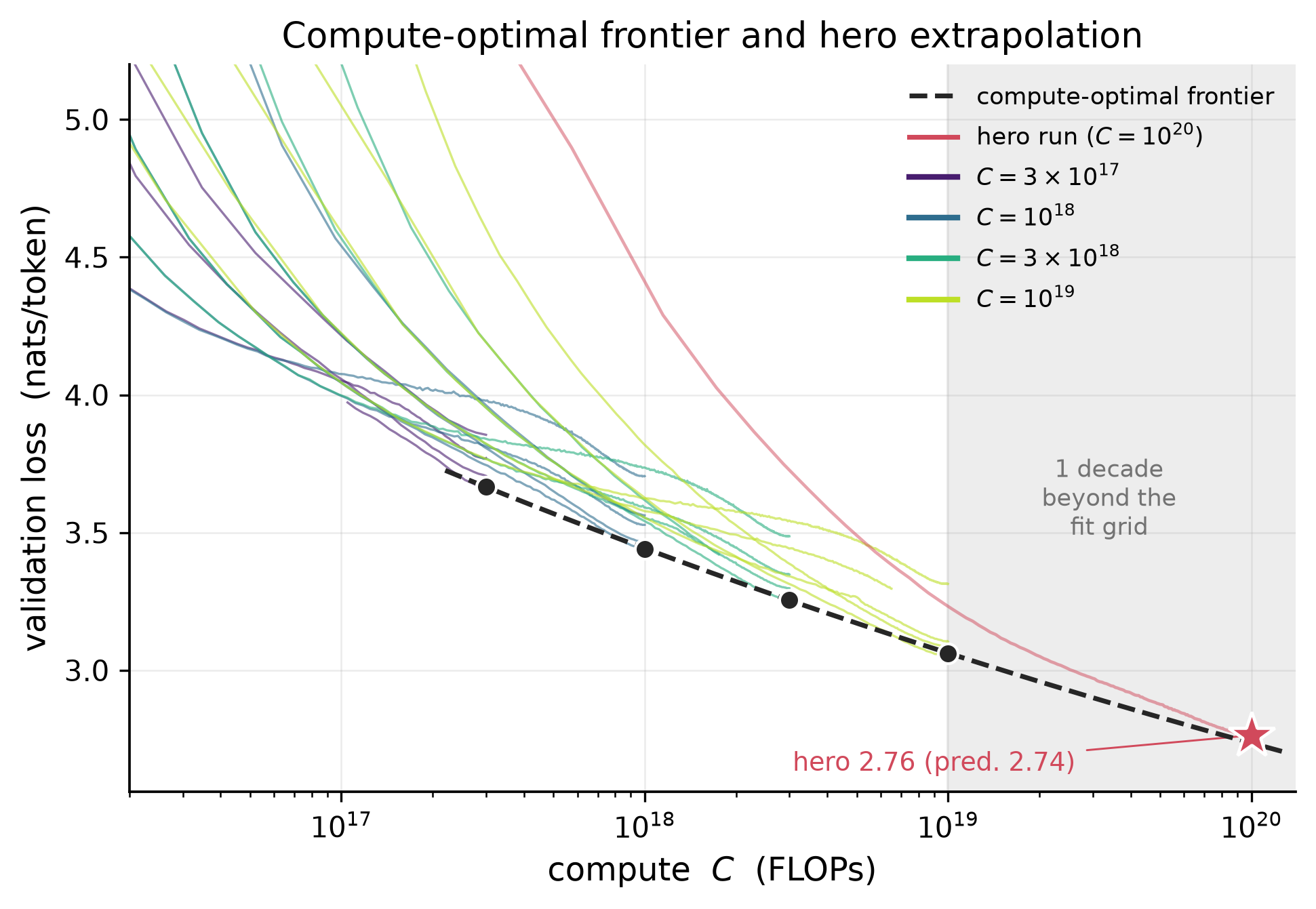}
        \caption{}\label{fig:isoflop-b}
    \end{subfigure}
    \caption{\textbf{(a)} IsoFLOP profiles: validation loss $L$ versus model size $N$ at each compute budget $C$, with an interior compute-optimal minimum per tier. \textbf{(b)} The same runs as learning curves (loss versus compute $C=6N_{\backslash E}D$, colored by budget); their lower envelope is the compute-optimal frontier $L_{\mathrm{opt}}(C)$, which extrapolates a decade beyond the fit grid to the out-of-sample hero run at $C=10^{20}$ (measured $2.76$ vs.\ predicted $2.74$ nats).}
    \label{fig:isoflop}
\end{figure}

Figure~\ref{fig:isoflop} shows the validation loss $L$ profiles across the grid; each model is trained single-epoch to the token budget its compute tier and width imply, and the full loss grid is tabulated in Appendix~\ref{app:grid}.

Every tier produces the U-shaped profile in $\log N$ that the IsoFLOP method assumes, with an interior optimum well inside the sampled widths at all four compute budgets. The parabolas flatten as compute grows (leading quadratic coefficient in $\ln N$ $0.327 \to 0.099$ from $3\times10^{17}$ to $1\times10^{19}$), so the vertex is less tightly determined at high compute; this dominates our exponent uncertainty (Section~\ref{sec:sensitivity}).

\subsection{Compute-Optimal Frontier}
\label{sec:extrap}
Taking each parabola vertex as $(N_{\mathrm{opt}}, L_{\mathrm{opt}})$ with $N_{\mathrm{opt}}$ the total parameter count (our primary convention, Section~\ref{sec:convention}) and $D_{\mathrm{opt}} = C/(6N_{\mathrm{opt}})$, log-log regression over the four compute budgets ($R^2 = 0.996$; per-tier vertices in Appendix~\ref{app:frontier}) yields
\[
N_{T,\mathrm{opt}}(C) = 1.82\times10^{-2}\, C^{0.556}, \qquad D_{\mathrm{opt}}(C) \propto C^{0.444}.
\]
So $\alpha \approx 0.56$, close to the $0.5$ that \citet{hoffmann2022trainingcomputeoptimallargelanguage} report for dense Transformers. Because the grid is defined by $C \equiv 6ND$, $\beta = 1-\alpha$ is not estimated separately.

\paragraph{Out-of-sample extrapolation.} We train one compute-optimal model a decade beyond the grid, at $C = 10^{20}$ FLOPs ($N_T \approx 2.44$B, width $2048$/depth $41$; $D \approx 7.1$B tokens single-epoch, same $\mathrm{LR}_{\mathrm{opt}}(d)$ rule). Extrapolating $L_{\mathrm{opt}}(C)$ from the grid vertices predicts $L^\star \approx 2.74$ nats; the run reaches $L \approx 2.76$ nats, within $0.03$, so the frontier holds out of sample.

\subsection{The Parameter-Counting Convention}
\label{sec:convention}
Refitting under all three conventions (Table~\ref{tab:conventions}), $\alpha$ moves from $0.75$ (strict non-embedding, near Kaplan's reported $0.73$) through $0.62$ (FLOP-matmul) to $0.56$ (total, near Chinchilla's $0.5$), spanning essentially the entire Kaplan--Chinchilla gap on a single dataset and architecture, at matched fit quality ($R^2 \approx 0.996$).

\begin{table}[t]
  \caption{Compute-optimal exponent $\alpha$ ($N_{\mathrm{opt}}\propto C^{\alpha}$) under three parameter-counting conventions, estimated in two ways: the IsoFLOP vertex fit (Sec.~\ref{sec:isoflop-profiles}) and a Chinchilla Approach~3 parametric fit $L = E + A/N^a + B/D^b$ (Appendix~\ref{app:approach3}). The two estimators agree within each convention. $D/N$ is the compute-optimal token-to-parameter ratio at the smallest and largest budget using IsoFLOP fit.}

  \vspace{1em}
  \label{tab:conventions}
  \centering
  \small
  \setlength{\tabcolsep}{4pt}
  \begin{tabular}{l c c c c}
    \toprule
    Convention & $N$ & $\alpha$ (IsoFLOP) & $\alpha$ (Approach 3) & $D/N$ ($3{\times}10^{17}{\to}10^{19}$) \\
    \midrule
    Strict non-embedding (Kaplan)    & $13ld^2$       & 0.748 & 0.745 & $25.1 \to 4.7$ \\
    FLOP-matmul (grid build)  & $13ld^2 + vd$  & 0.624 & 0.644 & $9.2 \to 4.1$ \\
    Total (Chinchilla, primary) & $13ld^2 + 2vd$ & 0.556 & 0.595 & $4.9 \to 3.5$ \\
    \bottomrule
  \end{tabular}
\end{table}

This has two consequences. First, small-scale IsoFLOP exponents should not be compared across studies without matching the counting convention. Second, our apparent departure from $\alpha = 0.5$ is largely a counting artifact: the larger non-embedding exponents reflect the small-scale bias \citet{pearce2024reconcilingkaplanchinchillascaling} predict, not an architectural difference.

\subsection{Sensitivity and Cross-Check}
\label{sec:sensitivity}
Details are in Appendix~\ref{app:sensitivity}. A leave-one-group-out jackknife gives $\alpha_{\mathrm{total}} = 0.556$ with range $[0.532, 0.615]$ (mean $0.565 \pm 0.053$), so the estimate does not depend on any single tier. Repeating four compute-optimal configurations across additional seeds gives run-to-run variance $\leq 0.003$ nats, which shifts the exponent by $<0.01$, so the estimate is not seed-limited. It is far more sensitive to \emph{how many} tiers are fit: two, three, and four tiers give $0.454$, $0.537$, and $0.556$, so more tiers are the main way to tighten it. As an independent check, a Chinchilla Approach~3 parametric fit ($L = E + A/N^a + B/D^b$, which does not impose $C = 6ND$) reproduces the per-convention exponents to within $0.04$ (Table~\ref{tab:conventions}) and estimates an irreducible loss $E \approx 2.7$ nats (Appendix~\ref{app:approach3}).

\subsection{Comparison to Non-Spiking Baselines}
\label{sec:tax}
\begin{figure}[t]
  \centering
  \includegraphics[width=0.85\textwidth]{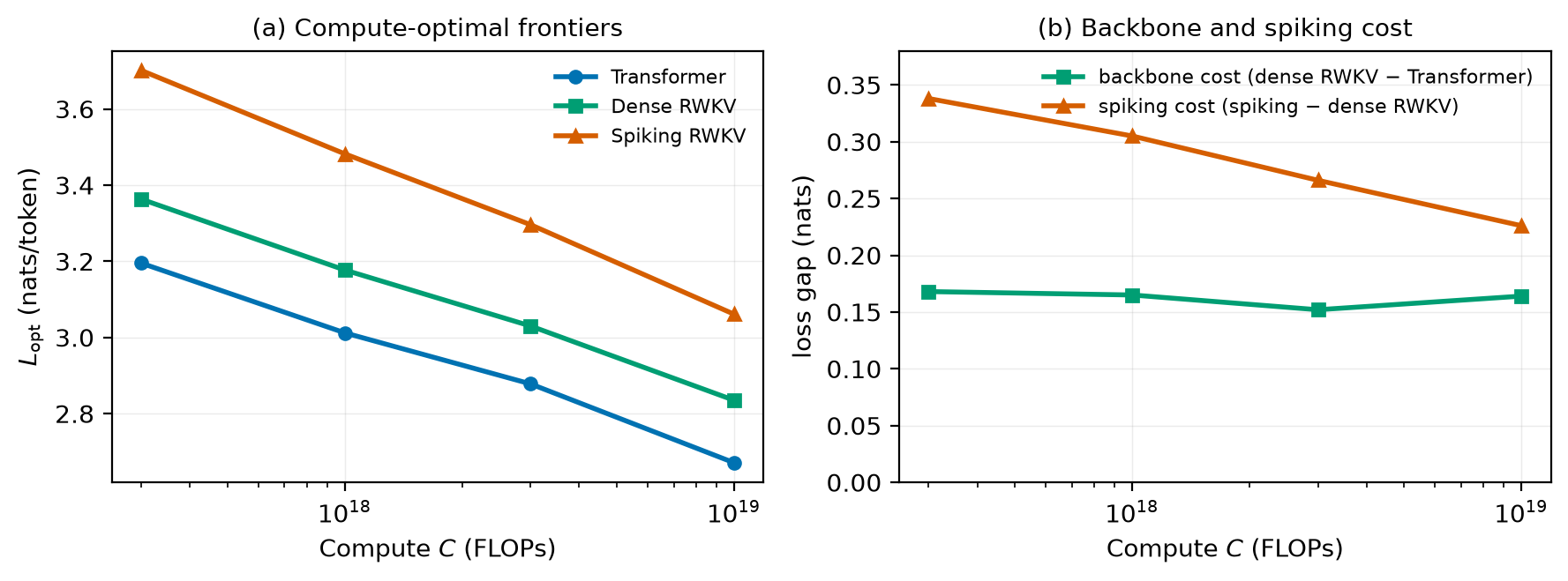}
  \caption{Spiking RWKV compared with non-spiking baselines. \textbf{(a)} Compute-optimal loss $L_{\mathrm{opt}}(C)$ for each architecture. \textbf{(b)} Backbone cost (dense RWKV minus Transformer) and spiking cost (spiking minus dense RWKV) at each budget.}
  \label{fig:triad}
\end{figure}

To measure what spiking costs in loss, we repeat the IsoFLOP grid for two non-spiking baselines, a continuous-valued RWKV and a standard softmax-attention Transformer, using the same data, batch size, learning-rate schedule, and evaluation as the spiking runs and tuning only the learning rate per architecture. Comparing compute-optimal losses at each budget separates two effects (Figure~\ref{fig:triad}): the backbone cost, the gap between dense RWKV and the Transformer, and the spiking cost, the gap between spiking and dense RWKV.

The backbone cost stays between $0.15$ and $0.17$ nats at every budget (Table~\ref{tab:triad}, Appendix~\ref{app:triad}). The spiking cost falls steadily, from $0.338$ to $0.305$, $0.266$, and $0.225$ nats between $3\times10^{17}$ and $10^{19}$ FLOPs, about $0.075$ nats per decade of compute, and is the only part of the gap that shrinks with scale. At $10^{19}$ we add a width-$2048$ dense RWKV run above the compute-optimal size, so its minimum is not at the edge of the sampled widths.

Finally, spike rates depend on training progress rather than model size, rising toward a $\sim$0.36 firing fraction as tokens-per-parameter grows (Appendix~\ref{app:sparsity}).

\section{Conclusion and Limitations}
\label{sec:limitations}
\label{sec:conclusion}
Under total-parameter counting, the compute-optimal frontier of a spiking language model is $N_{T,\mathrm{opt}} \propto C^{0.556}$, close to dense Transformers' $0.5$; larger non-embedding exponents come from parameter counting, not the architecture. About $0.16$ nats of the gap to a Transformer comes from the RWKV backbone and does not change with compute, while the cost of spiking falls steadily, from $0.34$ to $0.23$ nats between $3\times10^{17}$ and $10^{19}$ FLOPs.
These results come from four tiers between $3\times10^{17}$ and $10^{19}$ FLOPs, far below Chinchilla's $10^{21}$--$10^{24}$, on FineWeb-EDU only, with the aspect ratio $d/l$ drifting from $38$ to $53$ across widths. The baselines use one seed per configuration, so whether the spiking cost keeps falling at larger scale is untested; more tiers beyond the $10^{20}$-FLOP point (Section~\ref{sec:extrap}) would test both this and the power law.

\begin{ack}
We would like to thank Lambda for their generous computational grant through the Lambda Research Grant Program. We also thank Cooper Sigrist and Charlie Guo from X School for their support.
\end{ack}

\bibliographystyle{plainnat}
\bibliography{references}

\appendix

\section{Baseline Compute-Optimal Losses}
\label{app:triad}
\begin{table}[H]
  \caption{Compute-optimal loss $L_{\mathrm{opt}}$ (nats) for spiking RWKV and the two non-spiking baselines (Section~\ref{sec:tax}), with backbone cost (dense RWKV minus Transformer) and spiking cost (spiking minus dense RWKV). All runs share data, batch size, schedule, and evaluation; the learning rate is tuned per architecture. The $10^{19}$ dense RWKV fit includes a width-$2048$ run above the compute-optimal size.}
  \vspace{0.5em}
  \label{tab:triad}
  \centering\small
  \begin{tabular}{l c c c c c}
    \toprule
    $C$ (FLOPs) & Transformer & Dense RWKV & Spiking RWKV & backbone cost & spiking cost \\
    \midrule
    $3\times10^{17}$ & 3.195 & 3.363 & 3.701 & 0.168 & 0.338 \\
    $1\times10^{18}$ & 3.012 & 3.177 & 3.482 & 0.166 & 0.305 \\
    $3\times10^{18}$ & 2.878 & 3.030 & 3.296 & 0.151 & 0.266 \\
    $1\times10^{19}$ & 2.671 & 2.835 & 3.061 & 0.164 & 0.225 \\
    \bottomrule
  \end{tabular}
\end{table}

\section{Full Loss Grid}
\label{app:grid}
\begin{table}[H]
  \caption{Final validation loss across the IsoFLOP grid. $N$ is the total parameter count. $d$ is the embedding dimension. $l$ is the number of layers. Bold marks the per-tier minimum; dashes indicate configurations outside the grid at that budget.}
  \label{tab:profiles}
  \centering
  \small
  \begin{tabular}{lr rrrr}
    \toprule
    & & \multicolumn{4}{c}{Compute budget $C$ (FLOPs)} \\
    \cmidrule(lr){3-6}
    $d$/$l$ & $N$ & $3\times10^{17}$ & $1\times10^{18}$ & $3\times10^{18}$ & $1\times10^{19}$ \\
    \midrule
    384/10  & 57.8M   & 3.8049 & 3.7049 & -- & -- \\
    512/12  & 92.4M   & \textbf{3.7002} & 3.5283 & 3.4866 & -- \\
    640/14  & 138.9M  & 3.7384 & \textbf{3.4413} & 3.3480 & 3.3141 \\
    768/18  & 215.2M  & 3.8867 & 3.4621 & 3.2978 & 3.2030 \\
    1024/20 & 375.6M  & -- & 3.5627 & \textbf{3.2558} & 3.1041 \\
    1280/24 & 639.8M  & -- & -- & 3.3020 & \textbf{3.0506} \\
    1536/29 & 1043.8M & -- & -- & -- & 3.0838 \\
    \bottomrule
  \end{tabular}
\end{table}

\section{Compute-Optimal Frontier Vertices}
\label{app:frontier}
Table~\ref{tab:frontier} gives the per-tier compute-optimal allocation (total-parameter convention) recovered from the IsoFLOP parabola vertices, with $D_{\mathrm{opt}} = C/(6N_{\mathrm{opt}})$.
\begin{table}[H]
  \caption{Compute-optimal allocation from the IsoFLOP parabola vertices (total parameters).}
  \label{tab:frontier}
  \centering
  \small
  \begin{tabular}{l rrrr}
    \toprule
    $C$ (FLOPs) & $N_{\mathrm{opt}}$ & $D_{\mathrm{opt}}$ & $D/N$ & $L_{\mathrm{opt}}$ \\
    \midrule
    $3\times10^{17}$ & 101M  & 0.49B & 4.9 & 3.7014 \\
    $1\times10^{18}$ & 175M  & 0.95B & 5.5 & 3.4406 \\
    $3\times10^{18}$ & 349M  & 1.43B & 4.1 & 3.2564 \\
    $1\times10^{19}$ & 693M  & 2.40B & 3.5 & 3.0619 \\
    \bottomrule
  \end{tabular}
\end{table}

\section{Learning Rate Pilot}
\label{app:lr}
\begin{figure}[H]
    \centering
    \includegraphics[width=0.7\textwidth]{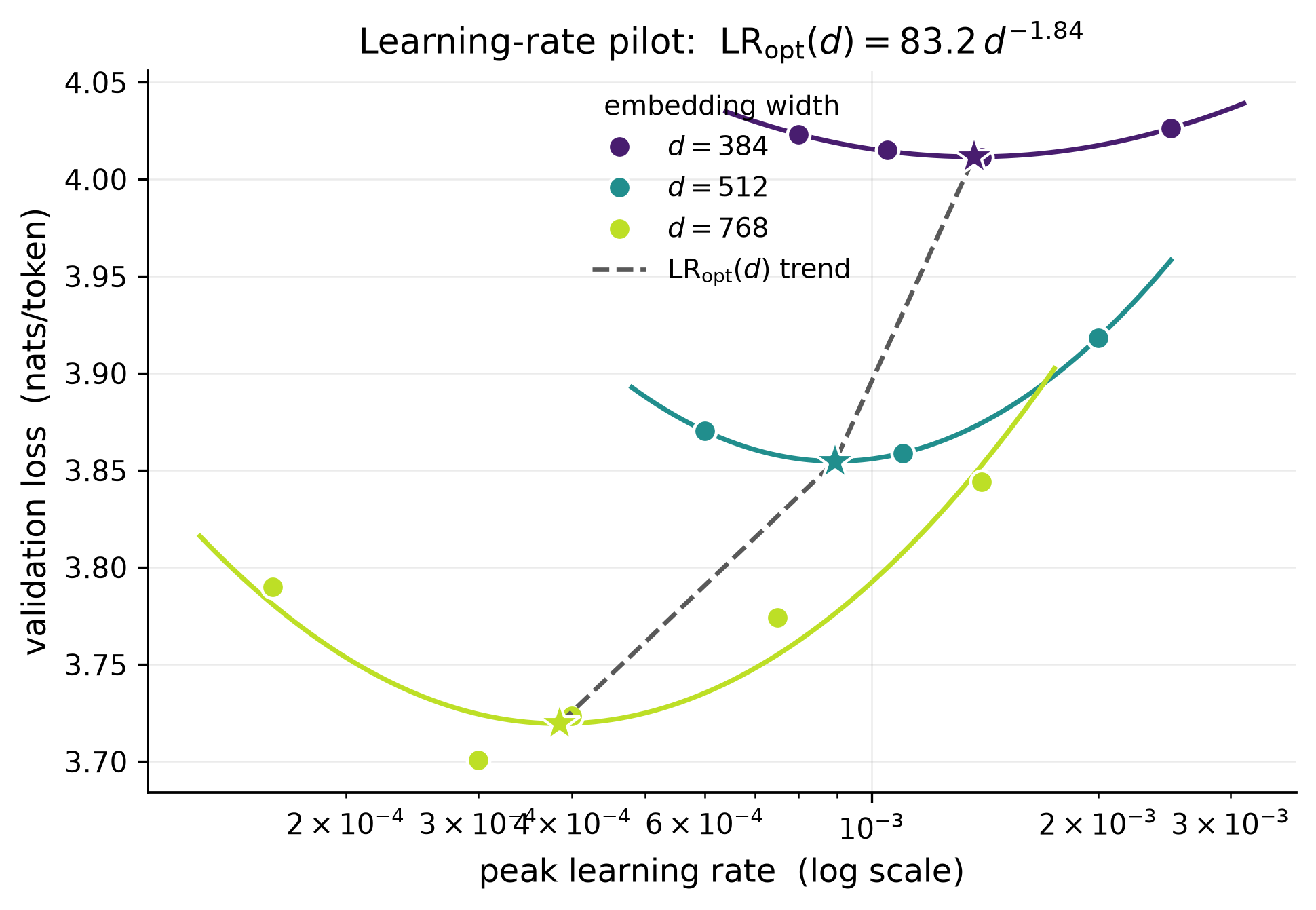}
    \caption{Fitted learning-rate power law for SpikeGPT.}
    \label{fig:lr}
\end{figure}

The optimal learning rate typically decreases with model width, with common rules $\mathrm{LR}_{\mathrm{opt}} \propto d^{-1}$ or $d^{-1/2}$ \citep{yang2022featurelearninginfinitewidthneural}, \citep{yang2022tensorprogramsvtuning}. We sweep the learning rate over embedding dimensions and fit $\mathrm{LR}_{\mathrm{opt}}(d) = k\,d^{-p}$ by log-log regression (nesting both rules as $p=1,\tfrac12$), then set the peak learning rate at each width in the main sweep, with cosine decay to one-tenth. Figure~\ref{fig:lr} shows the fit.

\section{Parameter-Count Inversion (Cardano)}
\label{app:cardano}
With aspect ratio $A = d/l$ where $d$ is the embedding dimension and $l$ is the number of layers, each block contributes $13d^2$ matmul parameters from its time-mix and channel-mix components, so $N_{\backslash E} = 13d^3/A + vd$ and $N_T = N_{\backslash E} + vd$ where $v$ is the vocabulary size. Inverting a target $N_{\backslash E}$ to a width $d$ requires solving the cubic
\[
13d^3 + Avd - AN_{\backslash E} = 0,
\]
which has a closed form via Cardano's formula. Because $N_{\backslash E}$ and $d$ are related by this cubic rather than a monomial, $d_{\mathrm{opt}}(C)$ is not itself a power law except in the large-$d$ limit where $13d^3$ dominates $Avd$; we therefore sample target values $N_{\backslash E}$ along the fitted frontier and invert numerically at each, rounding to architecture-compatible widths.

\section{Sensitivity Detail}
\label{app:sensitivity}
Table~\ref{tab:jackknife} reports the leave-one-group-out jackknife for both the FLOP-matmul exponent $\alpha$ and the total-parameter exponent $\alpha_{\mathrm{total}}$. Dropping \texttt{c3e17} (the lowest-compute tier) produces the largest excursion; jackknife means sit close to the full-data estimates, so no single tier dominates the fit. Replacing the parabola vertex with the discrete per-tier argmin gives $\alpha_{\mathrm{total}} = 0.584$ but is markedly less stable under leave-one-out, since the argmin moves only in the discrete jumps the width grid allows; we report the vertex fit as primary.
\begin{table}[H]
  \caption{Leave-one-group-out jackknife on the IsoFLOP tiers.}
  \label{tab:jackknife}
  \centering
  \small
  \begin{tabular}{l rr}
    \toprule
    Dropped tier & $\alpha$ & $\alpha_{\mathrm{total}}$ \\
    \midrule
    \texttt{c3e17} & 0.674 & 0.615 \\
    \texttt{c1e18} & 0.616 & 0.553 \\
    \texttt{c3e18} & 0.625 & 0.560 \\
    \texttt{c1e19} & 0.605 & 0.532 \\
    \midrule
    Full-data      & 0.624 & 0.556 \\
    Jackknife mean & $0.630 \pm 0.046$ & $0.565 \pm 0.053$ \\
    \bottomrule
  \end{tabular}
\end{table}

\section{Approach 3 Parametric Fit}
\label{app:approach3}
We fit $L(N,D) = E + A/N^{a} + B/D^{b}$ jointly across all runs; the frontier exponent follows as $\alpha = b/(a+b)$ and does not impose $C = 6ND$. Per convention it gives $0.745$ (strict non-embedding \citet{kaplan2020scalinglawsneurallanguage}), $0.644$ (FLOP-matmul), and $0.595$ (total), matching the IsoFLOP estimates to within $0.04$, and an irreducible loss $E \approx 2.6$--$2.8$ nats. A five-parameter surface on 19 points is degenerate, so we rely on the agreement of the two estimators rather than either point estimate.

\section{Activation Sparsity}
\label{app:sparsity}
\begin{figure}[H]
    \centering
    \includegraphics[width=0.7\textwidth]{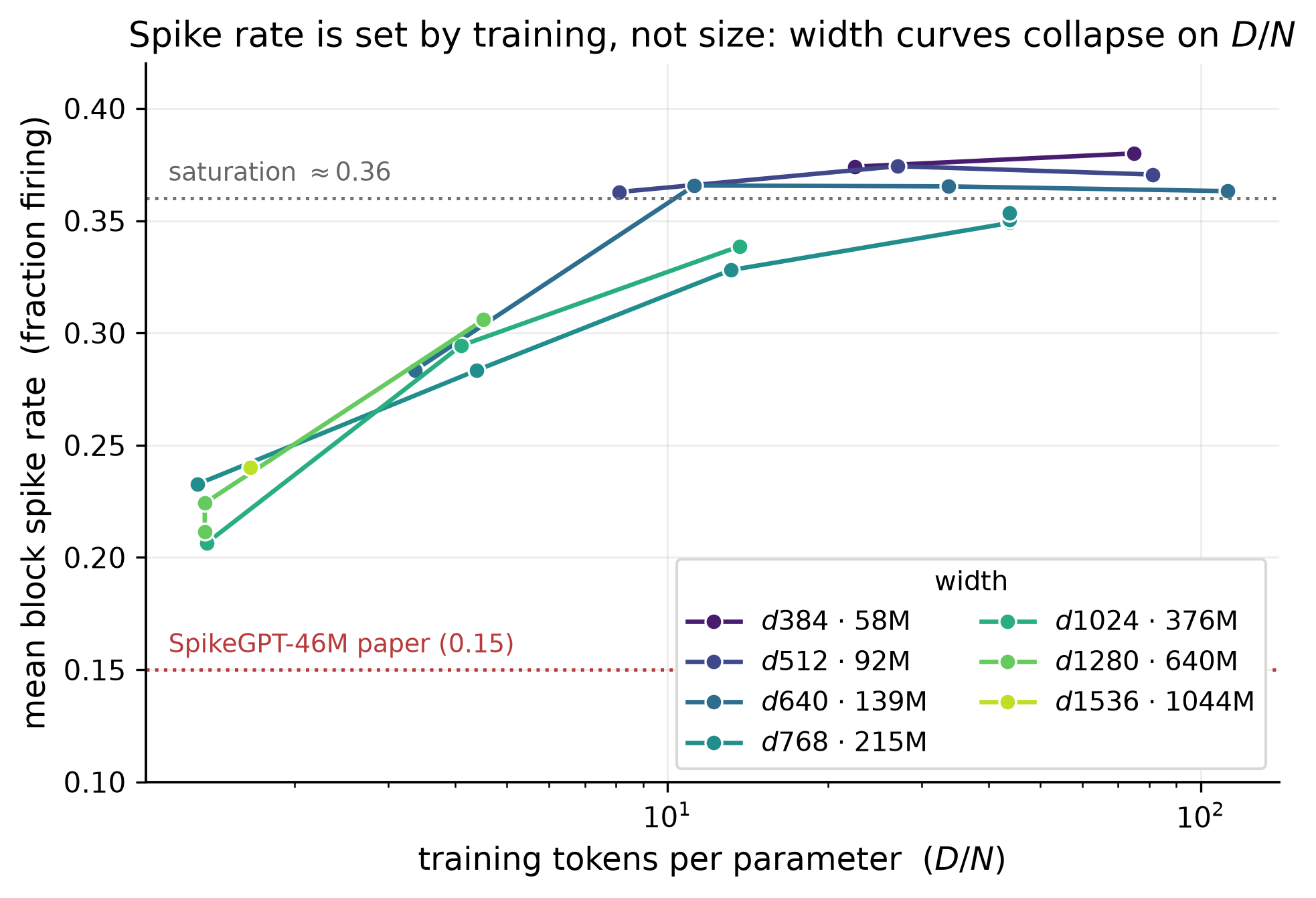}
    \caption{Mean block spike rate versus tokens per parameter; width curves collapse onto one trend.}
    \label{fig:sparsity}
\end{figure}

Figure~\ref{fig:sparsity} plots the mean block spike rate against training tokens per parameter where block spike rate is the fraction of neurons firing in a specific layer. The per-width curves collapse onto a single rising, saturating trend approaching a $\sim$0.36 firing fraction, indicating that sparsity is set by training progress rather than model size.

\end{document}